\documentclass{article}
\usepackage[linesnumbered,ruled,vlined]{algorithm2e}

\usepackage{arxiv}

\usepackage[utf8]{inputenc} 
\usepackage[T1]{fontenc}    
\usepackage{hyperref}       
\usepackage{url}            
\usepackage{booktabs}       
\usepackage{amsfonts}       
\usepackage{nicefrac}       
\usepackage{microtype}      
\usepackage{lipsum}
\usepackage{graphicx}
\graphicspath{ {./images/} }

\title{Unsupervised Image Segmentation using Mutual Mean-Teaching}

\author{
 Zhichao Wu \\
  CUHK\\
  \texttt{zhichaowu@cuhk.edu.hk} \\
   \And
 Lei Guo \\
  Xiamen University Xiamen\\
  \texttt{gl5121405@gmail.com} \\
  \And
 Hao Zhang \\
   Yunnan University\\
   \And
   Dan Xu \\
 Yunnan University\\
  \texttt{danxu@ynu.edu.cn} \\
}

\begin{document}
\maketitle
\begin{abstract}
Unsupervised image segmentation aims at assigning the pixels with similar feature into a same cluster without annotation, which is an important task in computer vision. Due to lack of prior knowledge, most of existing model usually need to be trained several times to obtain suitable results. To address this problem, we propose an unsupervised image segmentation model based on the Mutual Mean-Teaching (MMT) framework to produce more stable results. In addition, since the labels of pixels from two model are not matched, a label alignment algorithm based on the Hungarian algorithm is proposed to match the cluster labels. Experimental results demonstrate that the proposed model is able to segment various types of images and achieves better performance than the existing methods. 
\end{abstract}


\section{Introduction}
\hspace*{1em} Image segmentation is an important task in image processing, which assigns similar feature region into a same cluster. In computer vision, labeled dataset established by organization for image segmentation such as PASCAL VOC 2012\cite{pascal-voc-2012},BSD\cite{martin2001database},  have been applied to several fileds. 
 Encouraged by this, convolutional neural networks (CNNs)\cite{krizhevsky2012imagenet} \cite{simonyan2014very}  have been successfully applied to image segmentation for autonomous driving or augmented reality. The advantage of CNN-based classifier systems is that they do not require manually designed complexly image features and provide a fully automated classifier. These methods\cite{long2015fully} \cite{zheng2015conditional} \cite{badrinarayanan2017segnet} have been proven to be useful in supervised image segmentation.
 However, large amount of unlabeled images and videos has not been leveraged effectively. 
 
\hspace*{1em}  Recently, unsupervised image segmentation can be roughly divided into two categories, model-based method and learning-based method. K-Means\cite{krishna1999genetic} and Graph-based\cite{felzenszwalb2004efficient} were two popular model-based methods. Although they are fast and simplest, their performance are still not well.
Learning-based method refers to extract feature from deep neural network model.  Unsupervised domain adaptation (UDA)\cite{song2020unsupervised} \cite{zhang2019self} is typically proposed to adapt the model trained on the dataset with no identity annotations. In order to improve segmentation and labeling accuracy, researchers have expanded the basic UDA framework. \cite{kanezaki2018unsupervised} and   \cite{kim2020unsupervised} propose a novel end-to-end network of unsupervised image segmentation that consists of normalization and an argmax function for differentiable clustering. Their results show better accuracy than existing methods.   
  
\hspace*{1em}  However, previous studies on unsupervised image segmentation exit two important problems: Firstly, not stable. Model performance is largely dependent on random initialization of model parameters, which usually produces unstable results. They usually need to be trained several times to get the better result. Secondly, lack of limitation on the number of clusters.  It is very difficult for model to learn the number of clusters without prior knowledge. A hyperparameter, cluster number $k$, must be inputted into the model to limit the number of clusters after image segmentation.
  
\hspace*{1em}To address these problems, we introduced an unsupervised Mutual Mean-Teaching (MMT) framework to effectively perform image segmentation, which is inspired by \cite{ge2020mutual}. Specifically, for an arbitrary image input, cluster labels of pixels are obtained by training two same networks in unsupervised learning. These two collaborative networks also generate labels by network predictions for training each other. The labels generated in this way contain lots of noise. To avoid training error amplification, the temporally average model of each network is applied to produce reliable soft labels for supervising the other network in a collaborative training strategy. Furthermore, 
since the labels of pixels from two model are not matched, a label alignment algorithm based on the Hungarian algorithm is proposed to match the cluster labels. 

\hspace*{1em}   In this paper, we make the following contributions
   \begin{itemize}
  \item [1)] 
  MMT framework is introduced into unsupervised image segmentation task to produce stable segmentation result.
  \item [2)]
  	A label alignment algorithm based on the Hungarian algorithm is proposed to match the cluster labels.
  \item [3)]
  Experimental results demonstrated proposed method got the better accuracy than existing methods while maintaining a stable result. 
\end{itemize}

\section{Related Work}
\hspace*{1em} Image segmentation is the task of assigning every pixel to a class label. Recently, image segmentation using deep neural networks has made great breakthrough.
A fully convolutional network(FCN)\cite{long2015fully} has been proposed to train end-to-end network, and has outperformed conventional segmentation such as K-means\cite{krishna1999genetic} which leverage the clusters for improved classification scores is to do clustering on the feature vector. Unfortunately, the result of FCN has produced low spatial resolution and blurring effects. This problem was addressed by adding the CRF layer in the \cite{zheng2015conditional} maintaining the end-to-end architecture. However, there is a large amount of unlabeled images in real world. Unsupervised image segmentation learning is attracted a lot of researchers attention and several related methods have been proposed.     \cite{kanezaki2018unsupervised} and \cite{kim2020unsupervised} proposed joint learning approach learns optimal CNN parameters for the image pixel clustering. The  practicability of this method is that CNN filters are known to be effective for texture recognition and segmentation\cite{cimpoi2015deep} \cite{He_2019_ICCV}. But due to the random initialization of the filters, unstable results are obtained from their model. There are two different solutions to solve this problem. First, the parameters of CNN filters have been fixed, such as transfer learning\cite{oquab2014learning} \cite{cao2018partial} . The second solution is to create consistent training supervisions for labeled/unlabeled data via different models predictions, such as Teacher-student models\cite{laine2016temporal} \cite{zhang2018deep}. This paper choose the latter solution.  

\hspace*{1em}Recently, \cite{ge2020mutual} proposed Mutual Mean-Teaching (MMT) to learn better features from the target domain via off-line refined hard pseudo labels and on-line refined soft pseudo labels in an alternative training manner. The elegant work shows that the MMT structure can be used in unsupervised learning. However, the most time-consuming component in the MMT is that each iteration requires label alignment using k-means and the number of labels is fixed in original MMT structure.This disadvantage is not available for our task. So we select other label alignment method without fixed number of label instead of using k-means.
\cite{fan2012cluster}and \cite{zhou2006clusterer} use overlap similarity rate as the coefficient matrix, hungary and implicit enumeration algorithm as basic algorithms to solve label alignment problem. It is shown that the algorithm provides convergence at a rate which is eventually effective.

\section{Proposed Method}
\begin{figure*}
  \centering
  \begin{tabular}{@{}c@{}}
    \includegraphics[width=\linewidth]{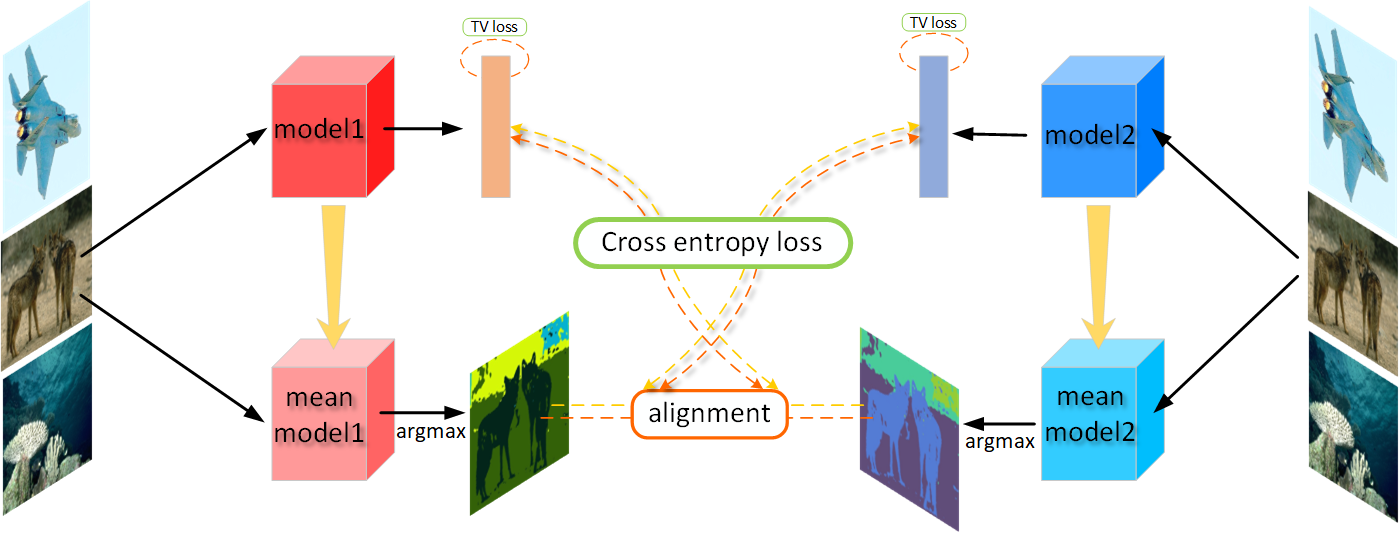} \\[\abovecaptionskip]
    \small (a) The pipeline of proposed model based MMT framework. 
  \end{tabular}

  \vspace{\floatsep}

  \begin{tabular}{@{}c@{}}
    \includegraphics[width=\linewidth]{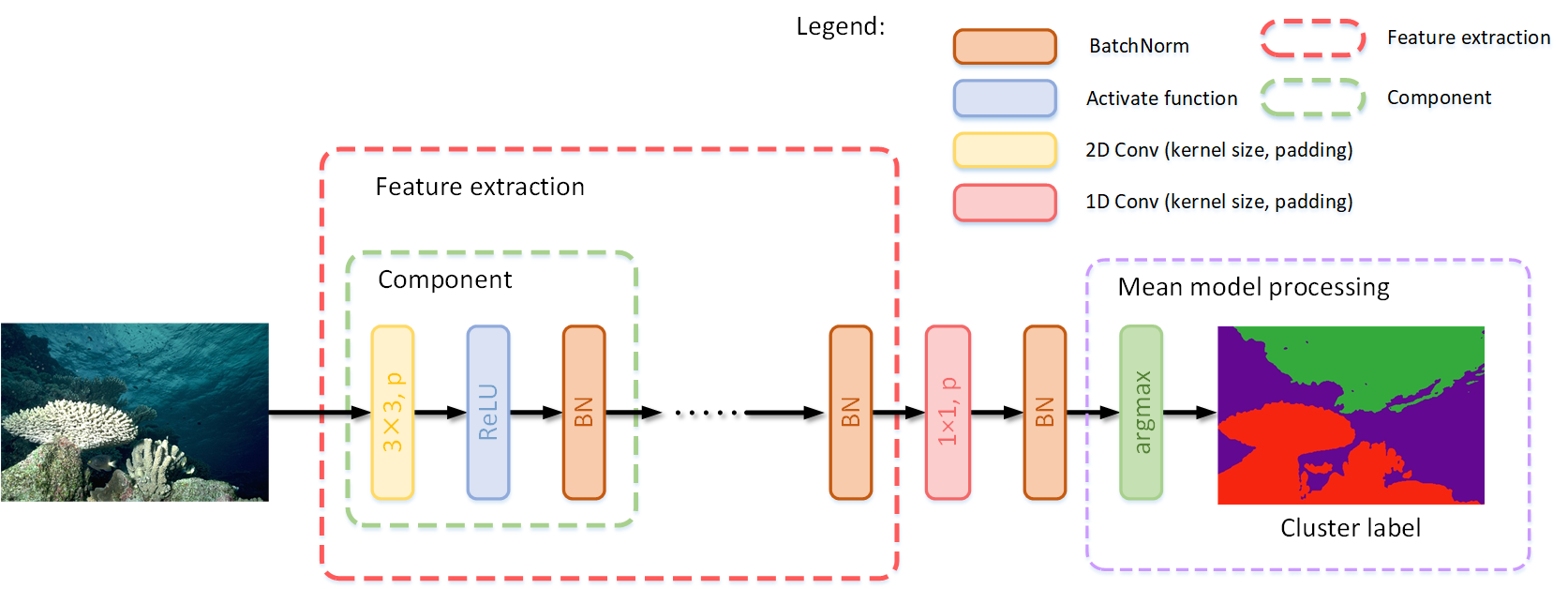} \\[\abovecaptionskip]
    \small  (b) The detail of model1 and model2.
  \end{tabular}

  \caption{Overall architecture of proposed model for unsupervised image segmentation.
(a) is the pipeline of proposed model based on MMT framework. It consists of two collaborative networks, model 1 and model 2. They adopt same structure which are optimized under supervisions of off-line refined labels and on-line refined labels.The loss function of total variation and label alignment cross entropy are adopted. (b) is details of model 1 and model 2. It consists of feature extraction module and mean model processing.  }\label{network}
\end{figure*}
\hspace*{1em} Image segmentation can be formulated as a classification problem of pixels with labels. For simplicity, let {} denote $\{\}^N_{n=1}$ unless otherwise noted, where $N$ denotes the number of pixels in an input color image $I=\{x_{n}  \in R^3\}$. Let $y_{n}=f(x_{n},\theta)$ simply denote the process of extracting feature, where $\theta$ denotes current network parameters, $f:R^3 \rightarrow R^q$ is a feature extraction function  and $\{y_{n} \in R^q \}$ be a set of $q$-dimensional feature vectors of image pixels. Cluster labels $\{c_{n} \in Z\}$ are assigned to all of pixels by $c_{n} = g(y_{n})$, where $g:R^{q} \rightarrow Z$ denotes a mapping function. $g$ can be an assignment function that return the label of the cluster centroid closest to $y_{n}$. The details interpretation are found in \cite{kim2020unsupervised}. A simple method coming our mind, while $f$ and $g$ are fixed, ${c_{n}}$ are obtained using the abovementioned process. Conversely, if $f$ and $g$ are trainable whereas ${c_{n}}$ are specified. then the parameters for $f$ and $g$ in this case can be optimized by gradient descent.
Although good results have been achieved in this architecture which meet the following three criteria: Pixels of similar features should be assigned the same
label; Spatially continuous pixels should be assigned the same
label;The number of unique cluster labels should be large.
However, the random initialization of filter parameters in model has crucial influences to the final performance.
So how to initialize the filter parameters randomly but also produce stable results ?

\hspace*{1em}we are motivated by the process of MMT creation: conduct label refinery in the target domain by optimizing the neural networks with off-line refined hard pseudo labels and on-line refined soft pseudo labels in a collaborative training manner. They propose this framework for tackling the problem of the label noise affecting the result performance significantly, especially in unsupervised learning. But the disadvantage is that there are fixed number of output label where it not available for our task requirements. So we select cluster label aligning algorithm \cite{zhou2006clusterer} without fixed number of label instead of using k-means. Our structure is proposed, as shown in Fig. 1. 





\subsection{Network architecture}
\hspace*{1em} In model1, given the input image $x_{n}$ where each pixel value is normalized. The network is trained to extract features transformation function $y_{n}^{1}=f_{1}(x_{n},\theta_{1})$.  Subsequently, to assign each pixel to a label is calculated by $C_{n}^{1} = argmax\{y_{n}^{1},  y_{n} \in R^q\}$. Similarly, model2 generates labels by the same networks with different initializations. We denote the feature transformation functions $y_{n}^{2}=f_{2}(x_{n},\theta_{2})$ and denote their corresponding label classifiers as $C_{n}^{2}$. If this two collaborative networks also generate  labels by network predictions for training each other, their labels  are lots of noise. In order to mitigate the label noise,  the past temporally average model of each network is applied to instead of the current model to generate the labels. Specifically, the parameters of the temporally average models of the two networks at current
iteration $T$ are denoted as $\theta_{mean1}^{T}$ and $\theta_{mean2}^{T}$ respectively, which can be calculated as

$$\theta_{1}^{T} = \theta_{1}^{T-1} +\eta\nabla \theta_{1} $$
$$\theta_{2}^{T} = \theta_{2}^{T-1}+\eta\nabla \theta_{2} $$
$$\theta_{mean1}^{T} = \alpha\theta_{mean1}^{T-1} + (1-\alpha)\theta_{1}^{T} $$
$$\theta_{mean2}^{T} = \alpha\theta_{mean2}^{T-1} + (1-\alpha)\theta_{2}^{T}$$

where $\theta_{mean1}^{T-1}$,$\theta_{mean2}^{T-1}$ indicate the temporal average parameters of the two networks in the
previous iteration $(T-1)$ ,the initial temporal average parameters are $\theta_{mean1}^{0} = \theta_{1}^{0}$ , $\theta_{mean2}^{0} = \theta_{2}^{0}$,$\alpha \in [0,1)$, $\eta$ is learning rate. 

\hspace*{1em}However, due to different initialization in the network, there is a great deal of difference between the two label generated by features of each channel in different network. As above assumption that  $C_{n}^{1} = \{c_{1}^{1}, c_{2}^{1},..., c_{n}^{1}\}$ and   $C_{n}^{2} = \{c_{1}^{2}, c_{2}^{2},..., c_{n}^{2}\}$, $C_{n}^{1}$ and $C_{n}^{2}$ is divided into $k$ sorts that can be expressed by the label vector $\lambda^{1} = [\lambda_{1}^{1},\lambda_{2}^{1},...,\lambda_{n}^{1}]$ and  $\lambda^{2} = [\lambda_{1}^{2},\lambda_{2}^{2},...,\lambda_{n}^{2}]$ where $\lambda_{i}^{1},\lambda_{i}^{2} \in \{1,2,...,k\}$  are the cluster label from different networks. In order to address the
issue of aligning label,  the overlap similarity matrix $S$ could be constructed, where $S_{i,j}$ items appear in both $\lambda_{i}^{1}$ and $\lambda_{j}^{2}$ is counted. Then, the pair of clusters whose number of overlapped data items is the largest, are matched in the way that they are denoted by the same label.Such a process is repeated until all the clusters are matched.The programming model is expressed as formula follows.
$$
\max Z =  sum \    S\bigodot D_{dec} = \sum_{i=1}^n\sum_{j=1}^n s_{i,j}\times d_{i,j} $$
$$
 s.t. \ \  \sum_{i=1}^n d_{i,j}  =  1(j=1,2,\dots,n)  \\ $$
 $$
       \sum_{j=1}^n d_{i,j} =    1(i=1,2,\dots,n) \\$$
       $$
      d_{i,j}  =  0 , 1 \\
$$
where $S$ is overlap similarity matrix, $D_{dec}$ is the decision-making matrix. $\bigodot$ is matrix points  multiplication sign.Briefly, we use $h$ denote the process of label alignment,

$$C_{n}^{2'} = h_{2}(C_{n}^{1})$$
$$ C_{n}^{1'} = h_{1}(C_{n}^{2})$$

where $h_{1}$ denotes the map from the space of $C_{n}^{2}$ to the space of $C_{1}$ and  $h_{2}$ denotes the map from the space of $C_{n}^{1}$ to the space of $C_{n}^{2}$ .
\begin{algorithm}[htb]

	\caption{Unsupervised image segmentation using MMT}
	\label{algoevent}
	\LinesNumbered
	\KwIn{I = {$x_{n} \in R^3$}}
	\KwOut{$L=\{c_{n} \in Z\}$t}
	$Init() \rightarrow \theta_{1},\theta_{2}$ \;
	$\theta_{1}^{mean1} = \theta_{1},  \theta_{2}^{mean2} = \theta_{2}$ \;

	\For{t=2:T}{
	    $y_{n}^{1}=GetFeats_{1}(x_{n})$\;
	    $y_{n}^{mean2}=GetFeats_{mean2}(x_{n})$\;
        $C_{n}^{mean2} = argmax\{y_{n}^{mean2}\}$\;
        $C_{n}^{1'} = h_{1}(C_{n}^{mean2})$ \quad  // \quad${Label Assignment}$\;
        $L_{sim}^{1} = L_{sim}(y_{n}^{1},C_{n}^{1'})$, $L_{tv}^{1} = L_{tv}(y_{n}^{1})$\;
        $L^{1} = L_{sim}^{1} + \beta L_{tv}^{1}$    \;
        $\theta_{1}^{t} = \theta_{1}^{t-1} +\eta \nabla \theta_{1}$, $\theta_{1,mean}^{t} = \alpha \theta_{1,mean}^{t-1} + (1-\alpha)\theta_{1}^{t}$   \;
        $y_{n}^{2} = GetFeats_{2}(x_{n})$\;
        $y_{n}^{mean1} = GetFeats_{mean1}(x_{n})$ \;
        $C_{n}^{mean1} = argmax\{y_{n}^{mean1}\}$, \;
         $C_{n}^{2'} = h_{2}(C_{n}^{mean1})$ \quad  // \quad${Label Assignment}$\;
        $L_{sim}^{2} = L_{sim}(y_{n}^{2},c_{n}^{2'})$, $L_{tv}^{2} = L_{tv}(y_{n}^{2})$ \; 
        $L^{2} = L_{sim}^{2} + \beta L_{tv}^{2}$\;
        $\theta_{2}^{t} = \theta_{2}^{t-1} +\eta\nabla \theta_{2}$ , $\theta_{mean2}^{t} = \alpha \theta_{mean2}^{t-1} + (1-\alpha)\theta_{2}^{t}$\;
		}

\end{algorithm}
\subsection{Loss Functions}
\hspace*{1em} \textbf{TV Loss}. To encourage the cluster labels to be the same as those of the neighboring
pixels. total variation regularizer would be considered to  horizontal and vertical differences of  output label.we follow prior work \cite{shibata2017misalignment}. The spatial continuity loss $L_{tv}$ is defined as follows:

$$
L_{tv}(y_{n}) = \sum_{i=1}^{W-1}\sum_{j=1}^{H-1}(||y_{i+1,j}-y_{i,j}||_{1} + ||y_{i,j+1}-y_{i,j}||_{1})
$$
where $W$ and $H$ represent the width and height of an input image, $|| \cdot ||_{1}$ represent the $L1-norm$.  whereas $y_{i,j}$ represents the pixel value at $(i,j)$ in the
feature map $y$.

\hspace*{1em}\textbf{Feature Loss}. the cluster labels generated by average model2 are obtained by applying the $argmax$ function to the feature map $y_{mean2}$. The cluster labels $c^{mean2}_{n}$are further utilized as the pseudo targets. And then align label with model1 by mapping function of $h_{1}$. The feature loss penalizes the output label when it deviates in content from the target $C_{n}^{mean2}$. To achieve
this goal, the following cross entropy loss is calculated:
$$
L_{sim}(y_{n},h_{1}(C^{mean2}_{n})) = \\ \sum_{n=1}^{N}\sum_{j=1}^{q}(-\delta(j-h_{1}(C^{mean2}_{n}))\ln{y}^{n,j}) 
$$
where 

$$ \delta(t)=\left\{
\begin{array}{rcl}
1       &      & {if \quad  t = 0}\\
0     &      & {otherwise}
\end{array} \right. $$

\subsection{OVERALL LOSS AND ALGORITHM}
\hspace*{1em} To finish the task of image segmentation, we assign label  to  match the
feature content representation of $y_{n}$ and spatial smoothness simultaneously. Thus we jointly minimise the distance of the content feature representations and spatial smoothness. The loss function we minimise is

$$L_{total} = L_{sim}+\beta L_{tv}$$

where $\beta$ is the weighting factor for spatial smoothness.

\hspace*{1em}The detailed optimization procedures are summarized in Algorithm 1. Compared to traditional unsupervised algorithm, labels generated by off-line refined instead of the labels generated by  on-line algorithm. We keep the model2 and mean model2 constant during the model1 training phase, As model1 training tries to figure out how to segment the image using  labels generated by mean model2, and then align the labels of the output of model1  and model2. Immediately, the parameter of $\theta_{1} $ and $\theta_{mean1}$ are updated.  Similarly, keep the model1 and mean model1 constant during the model2 training phase. the model2 would be trying to segment image according to  labels generated by mean model1. Then the parameter of  $\theta_{2} $ and $\theta_{mean2}$ are updated. After several steps of training, if model1,mean model1, model2 and mean model2 have enough capacity, they will reach a point at which four models are stable.






\section{EXPERIMENTAL RESULTS}
\begin{figure}[h]
\centering
\{

\centering
\includegraphics[width=0.71in]{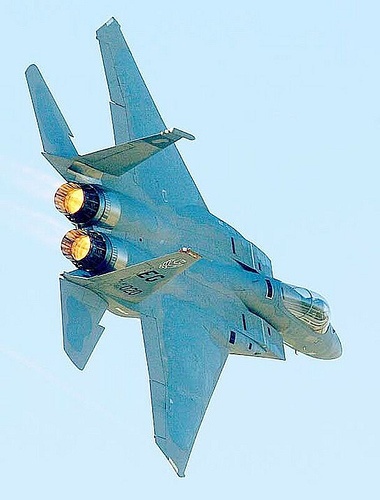}\vspace{1pt}
\includegraphics[width=0.71in]{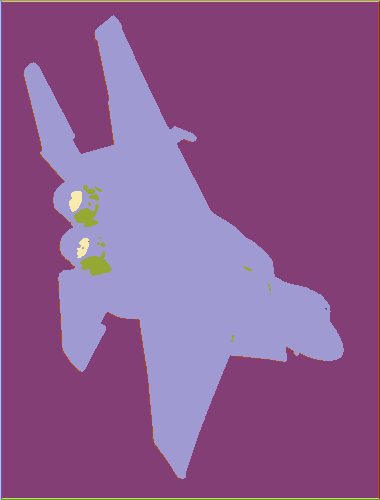}\vspace{1pt}
\includegraphics[width=0.71in]{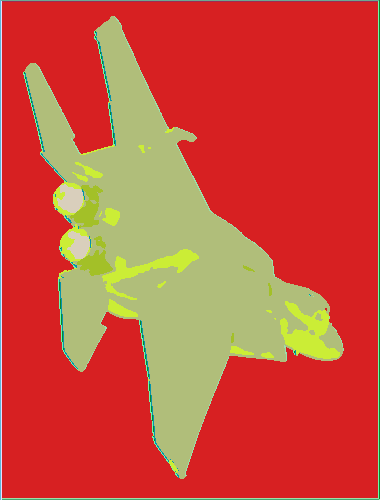}\vspace{1pt}
\includegraphics[width=0.71in]{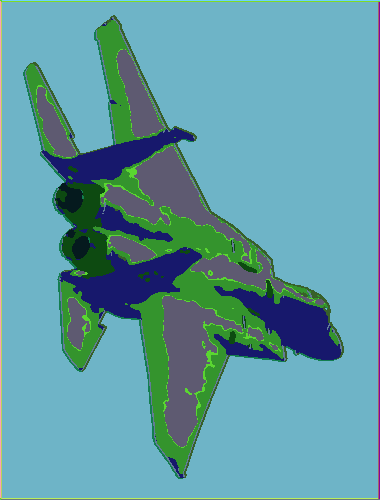}

\centering
\includegraphics[width=0.71in]{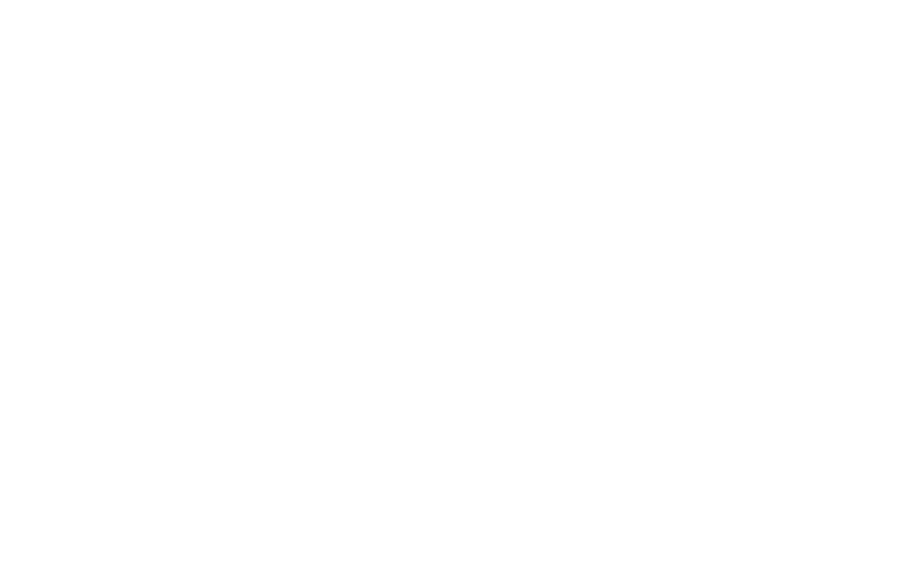}\vspace{1pt}
\includegraphics[width=0.71in]{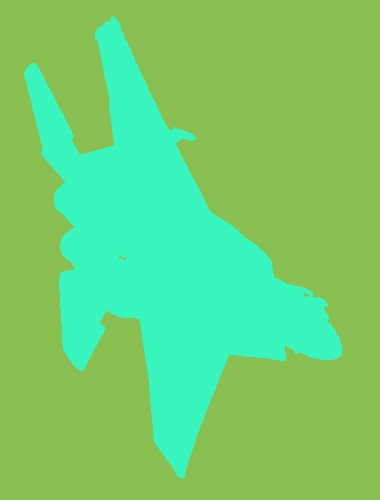}\vspace{1pt}
\includegraphics[width=0.71in]{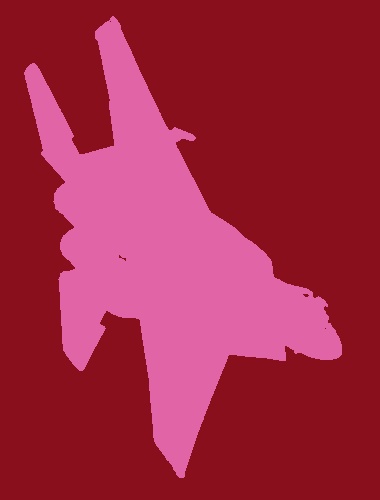}\vspace{1pt}
\includegraphics[width=0.71in]{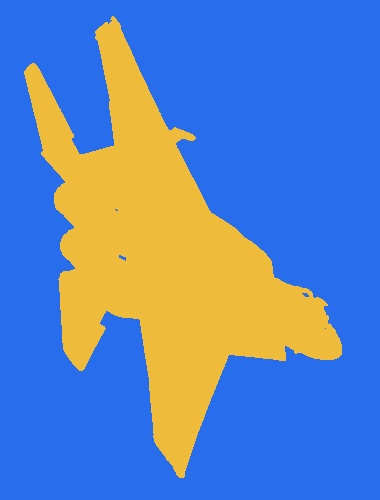}

\centering
\caption{Results of proposed method and \cite{kanezaki2018unsupervised} generated by training three times without fine-tuning any parameters. In first row, the first image is input image. The other three images are image segmentation results with different segments showing in different colors. The second row show three images are generated  by our method.}
\label{fig2}
\end{figure}

\begin{figure*}[htb]
\centering

\centering
img
\includegraphics[width=1.3in]{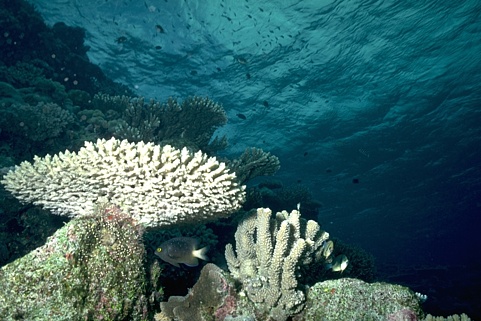}
\includegraphics[width=1.3in]{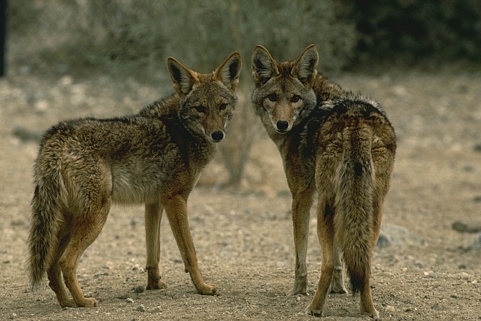}
\includegraphics[width=1.3in]{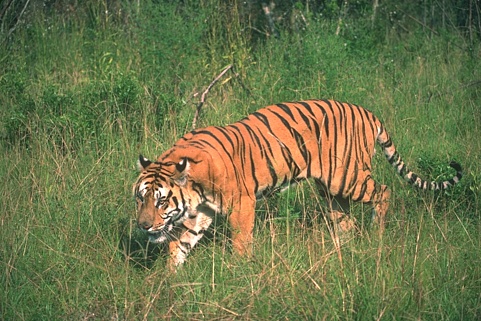}
\includegraphics[width=1.3in]{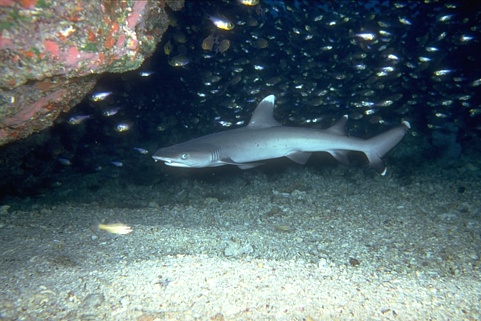}
\includegraphics[width=0.6in]{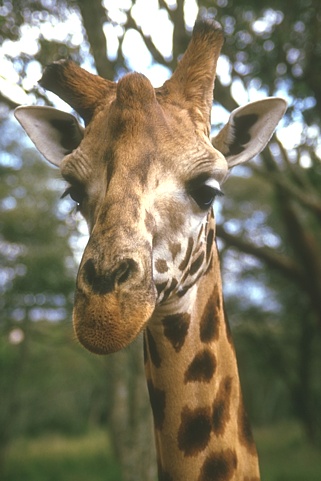}

\centering
gt\ \ \ \ 
\includegraphics[width=1.3in]{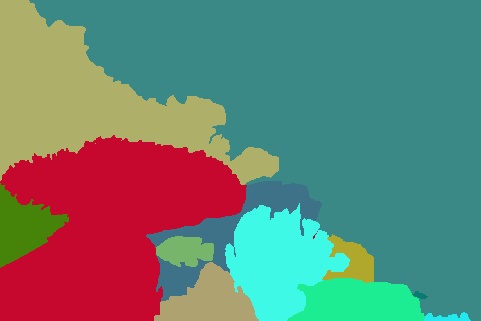}
\includegraphics[width=1.3in]{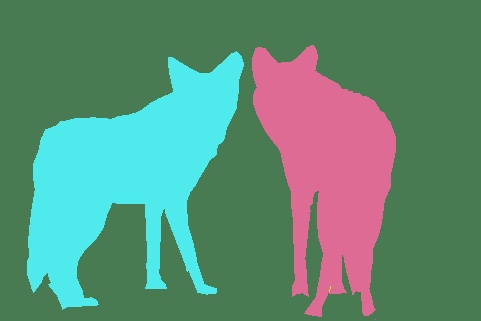}
\includegraphics[width=1.3in]{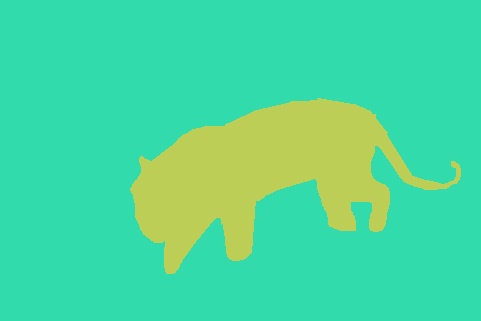}
\includegraphics[width=1.3in]{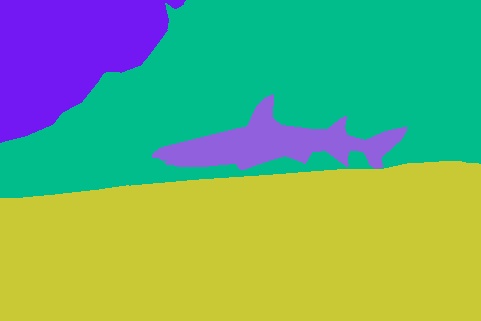}
\includegraphics[width=0.6in]{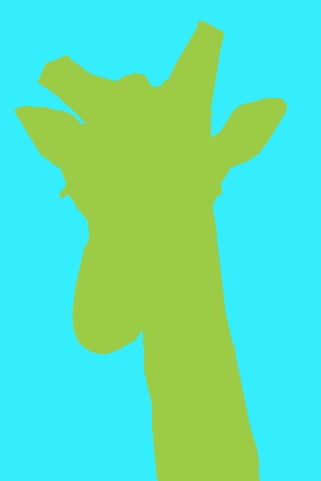}

\centering
\cite{kanezaki2018unsupervised}
\includegraphics[width=1.3in]{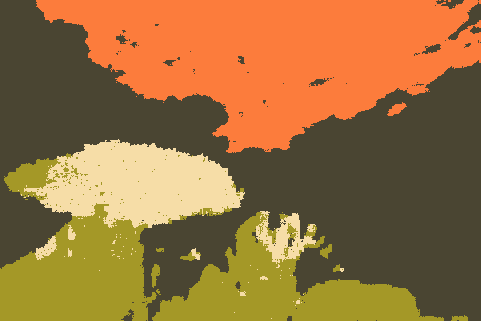}
\includegraphics[width=1.3in]{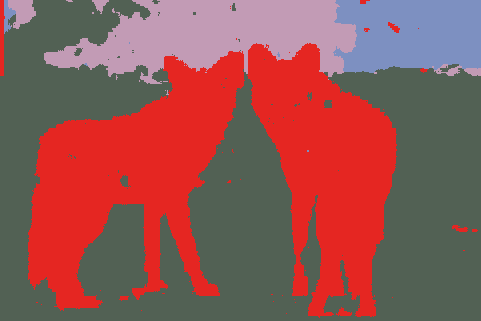}
\includegraphics[width=1.3in]{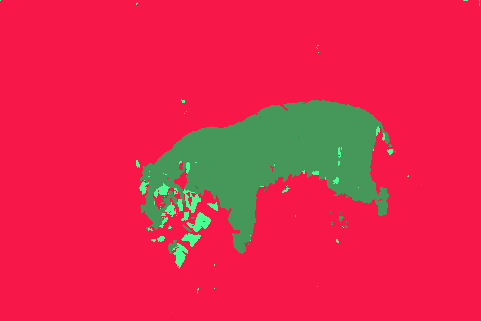}
\includegraphics[width=1.3in]{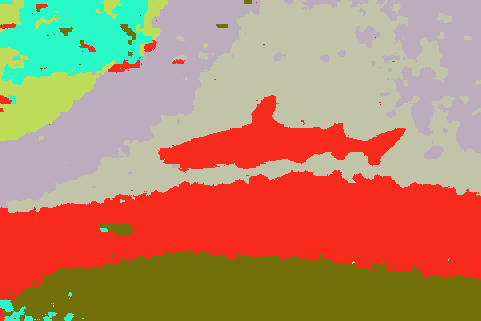}
\includegraphics[width=0.6in]{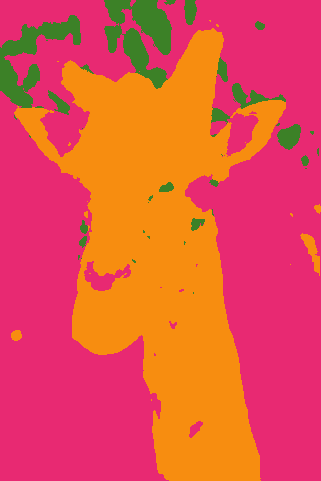}

\centering
ours
\includegraphics[width=1.3in]{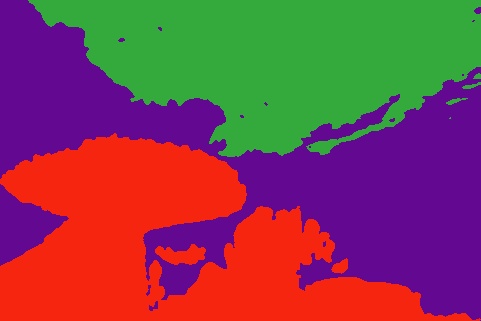}
\includegraphics[width=1.3in]{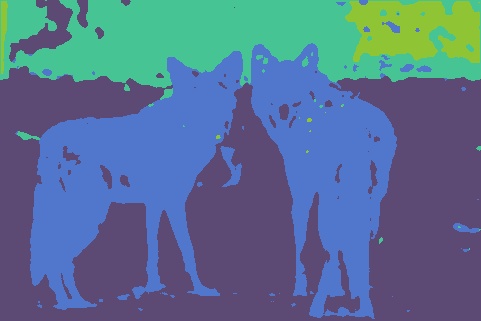}
\includegraphics[width=1.3in]{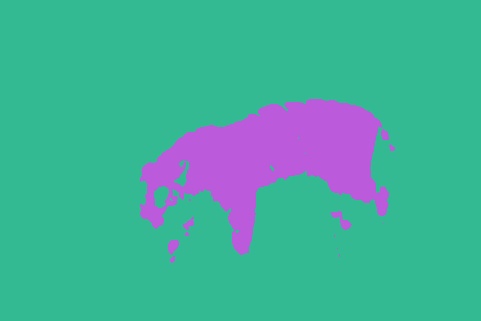}
\includegraphics[width=1.3in]{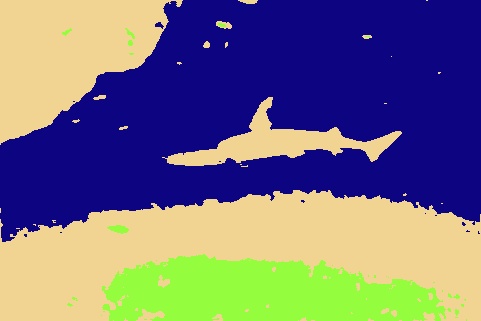}
\includegraphics[width=0.6in]{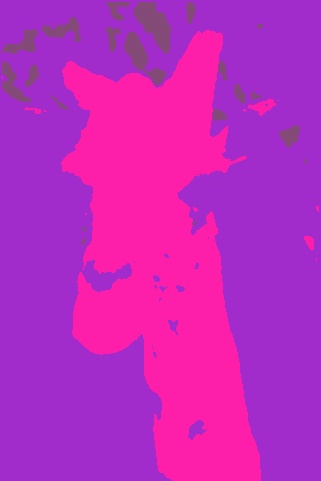}
\caption{ Comparison of our method and \cite{kanezaki2018unsupervised} on unsupervised image segmentation. The first two rows show the origin images and ground truth image segmentations.The third row show good results which are selected by training many times from \cite{kanezaki2018unsupervised}. In this process, failed results have been abandoned since their method is not stable. The last row is our proposed method result which is not training many times.  Different segments are shown in different colors.}
\label{fig1}
\end{figure*}

\hspace*{1em} In this section, we conduct experiments to verify the effectiveness
of the proposed stable image segmentation approach in unsupervised learning.
200 test images from the Benchmark(BSD500) are chosen to evaluate the proposed method. We trained the proposed model with T = 1000 iterations and fixed q = 100 for the channel of output feature map in the experiments. The number of convolutional layers M was set to 3 and the temporal ensemble momentum $\alpha$ is set to 0.999. The model parameters are optimized by the stochastic gradient descent method (SGD). The learning rate is fixed to 0.1 and set momentum to 0.9.

For the comparison, we reproduce the baseline image segmentation from \cite{kanezaki2018unsupervised}.  Figure \ref{fig2} shows the better results from our structure. The first row  which we repeat training 3 times without fine-tuning any parameters from  \cite{kim2020unsupervised} resulted in poor stable performance. Our method has a significant stable performance improvement.

\begin{figure*}[htb]
\centering

img
\includegraphics[width=0.58in]{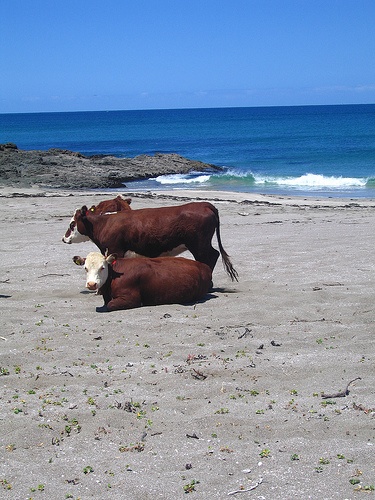}
\includegraphics[width=1.03in]{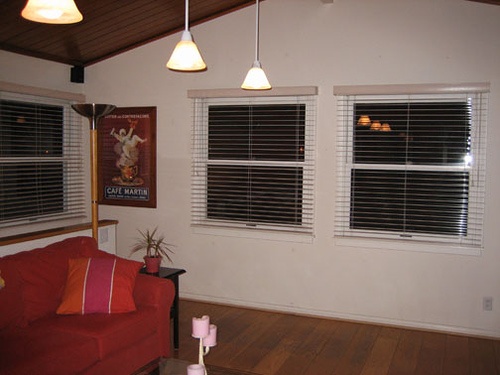}
\includegraphics[width=1.13in]{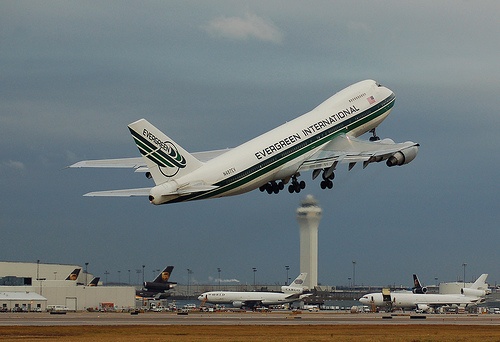}
\includegraphics[width=0.59in]{2010.jpg}
\includegraphics[width=0.9in]{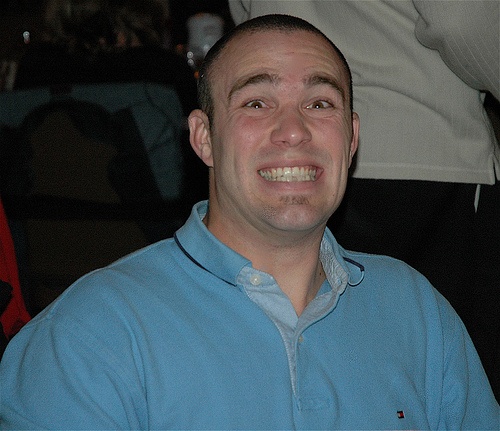}
\includegraphics[width=1.17in]{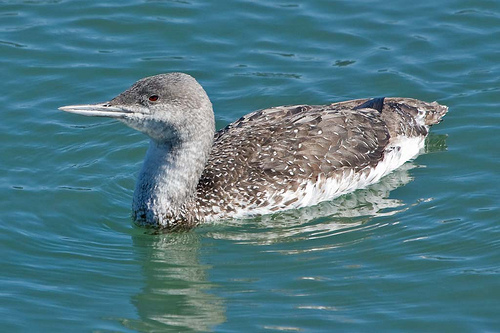}

\centering
gt\ \ \ \ \ 
\includegraphics[width=0.58in]{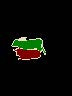}
\includegraphics[width=1.08in]{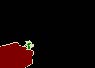}
\includegraphics[width=1.13in]{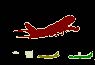}
\includegraphics[width=0.59in]{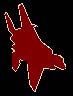}
\includegraphics[width=0.9in]{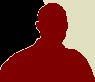}
\includegraphics[width=1.17in]{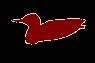}

\centering
\cite{ji2019invariant}
\includegraphics[width=0.58in]{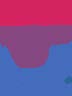}
\includegraphics[width=1.05in]{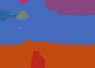}
\includegraphics[width=1.13in]{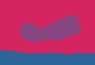}
\includegraphics[width=0.59in]{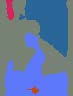}
\includegraphics[width=0.9in]{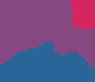}
\includegraphics[width=1.17in]{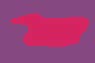}

\centering
\cite{kanezaki2018unsupervised}
\includegraphics[width=0.58in]{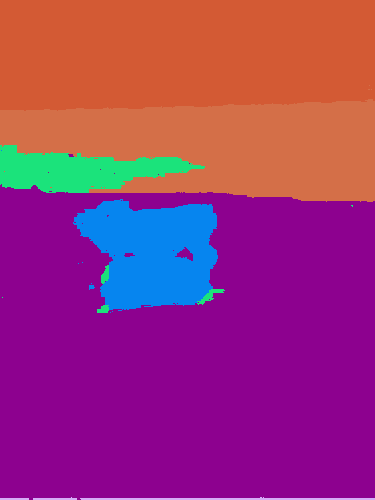}
\includegraphics[width=1.03in]{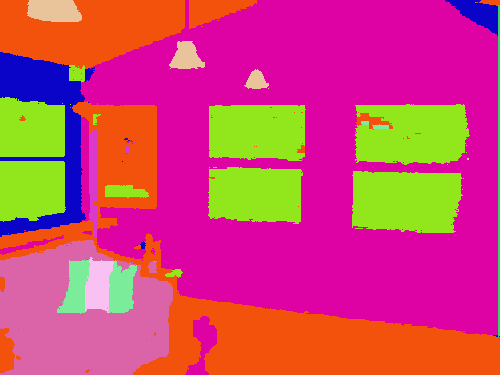}
\includegraphics[width=1.13in]{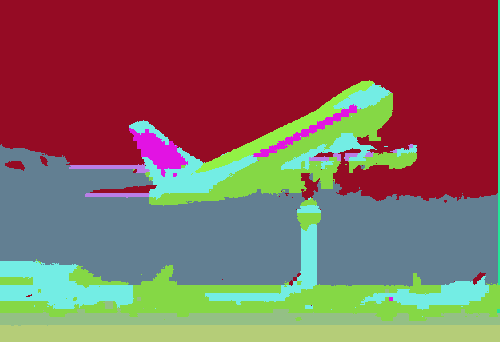}
\includegraphics[width=0.59in]{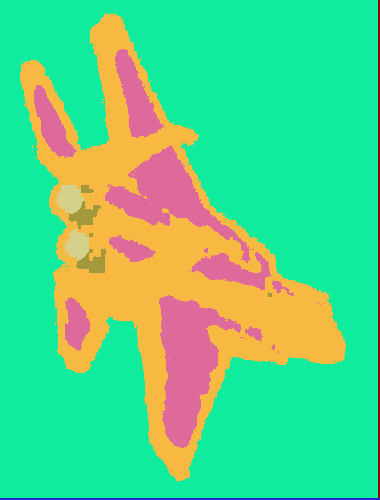}
\includegraphics[width=0.9in]{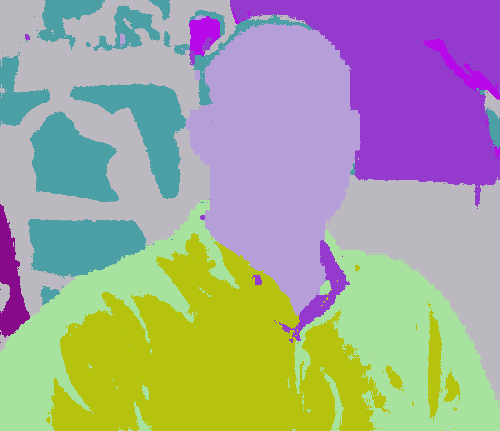}
\includegraphics[width=1.17in]{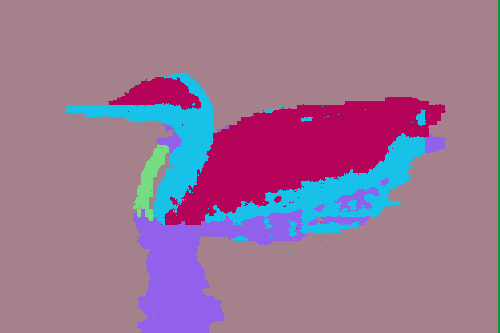}

\centering
\cite{kim2020unsupervised}
\includegraphics[width=0.58in]{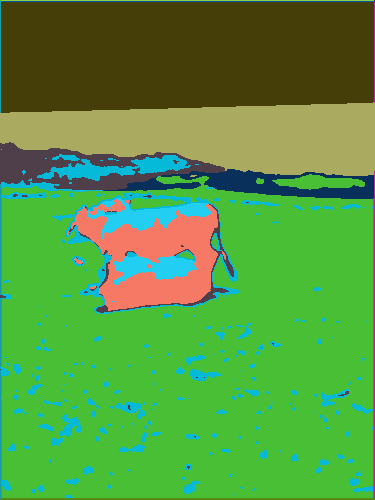}
\includegraphics[width=1.03in]{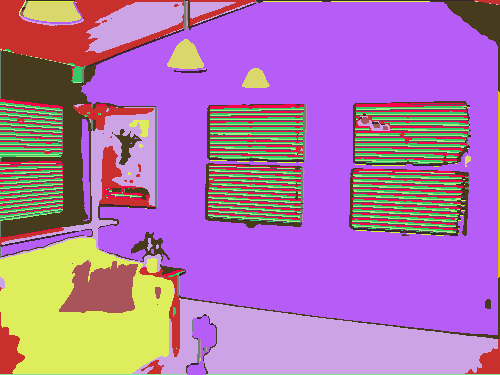}
\includegraphics[width=1.13in]{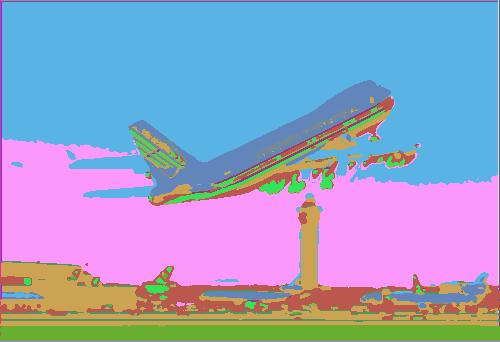}
\includegraphics[width=0.59in]{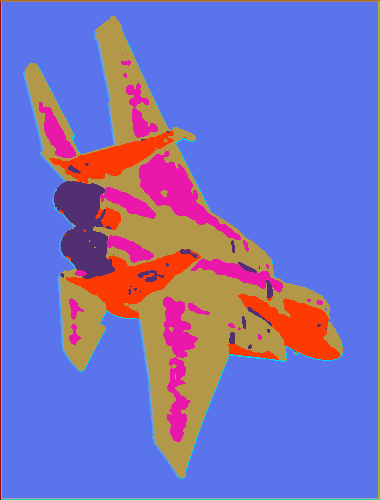}
\includegraphics[width=0.9in]{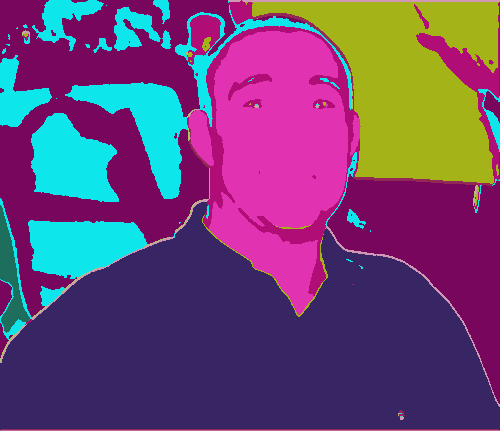}
\includegraphics[width=1.17in]{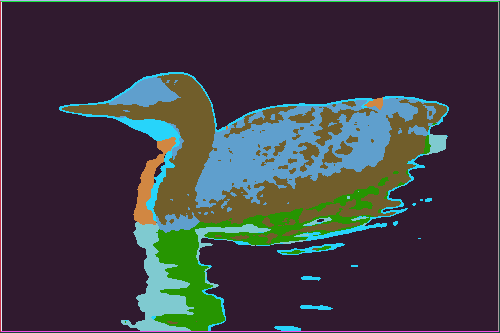}

\centering
ours
\includegraphics[width=0.58in]{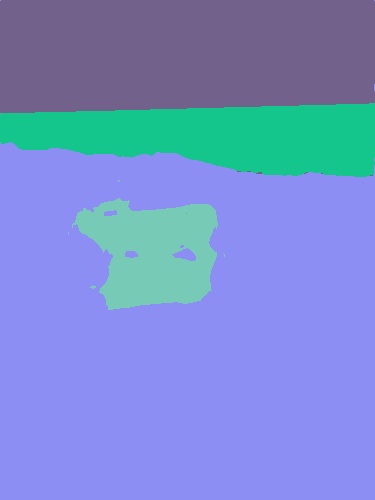}
\includegraphics[width=1.03in]{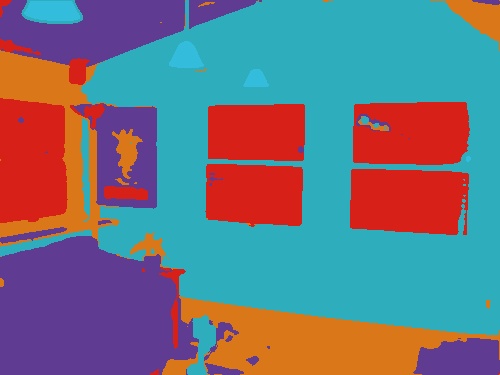}
\includegraphics[width=1.13in]{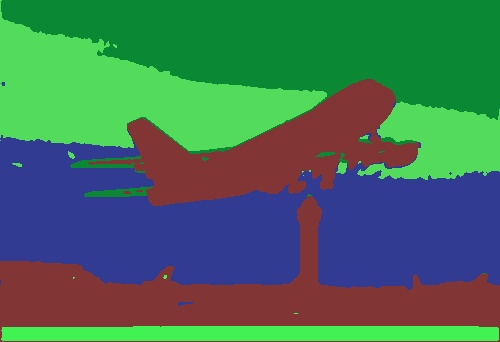}
\includegraphics[width=0.59in]{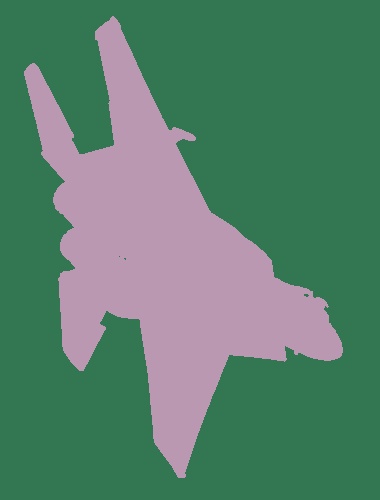}
\includegraphics[width=0.9in]{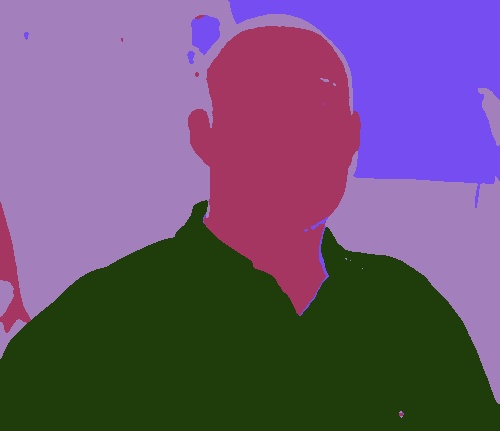}
\includegraphics[width=1.17in]{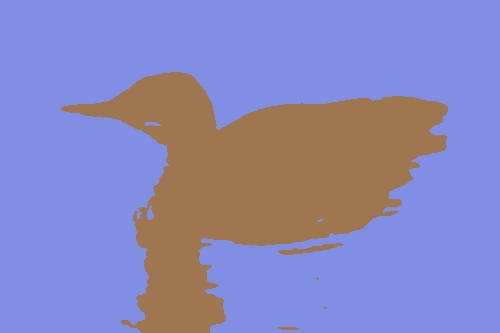}

\centering
\caption{Comparison of our method and IIC\cite{ji2019invariant}, \cite{kanezaki2018unsupervised},\cite{kim2020unsupervised} model fitting for six images from PAS-CAL VOC 2012. The first two rows show the origin images and ground truth image segmentations. The third row is generated by IIC method. The second and third shows
the segmentations produced by the previous state-of-the-art system
by \cite{kanezaki2018unsupervised} and \cite{kim2020unsupervised} and their failed results have been abandoned since their method is not stable. The last row shows the output of our highest performing structure.
}
\label{fig3}
\end{figure*}

\hspace*{1em}Examples of unsupervised image segmentation results on
PASCAL VOC 2012 and BSD500 are shown in Figure \ref{fig1} and
Figure \ref{fig3}, respectively. 
Figure \ref{fig1} shows a comparison of five images from
the BSD500. The first two rows show the origin images and ground
truth image segmentations, and third row shows the predicted image segmentation results with method of \cite{kanezaki2018unsupervised}, which I pick out relatively good results with training many times , and the last row are our results.They also demonstrate that our method performed better and  more consistently, which indicates that our model was able to learn features that are stable to a large range of intensity variations.The results of segmentation comparing our method with IIC\cite{ji2019invariant}, \cite{kanezaki2018unsupervised} and \cite{kim2020unsupervised}
 are shown in Figure \ref{fig3}. Note that our results in very smooth image segmentation which is closer to ground truth segmentation. The images in Figure \ref{fig3} offer compelling evidence that our segmentation algorithm performs well on a variety of images from different domains.

\hspace*{1em}The evaluation of segmentation results is that we can compare
the segmentations produced by different algorithms,
such as the k-means clustering, graph-based segmentation method (GS)\cite{felzenszwalb2004efficient} , \cite{kanezaki2018unsupervised} and \cite{kim2020unsupervised}. The parameters of \cite{kanezaki2018unsupervised} and \cite{kim2020unsupervised} are same as the parameter of we proposed method.The best parameters  of k-means clustering and graph-based segmentation algorithms were experimentally determined from $\{2,5,8,11,14,17,20 \}$ and $\{100,500,1000,1500,2000 \}$.
we get a accuracy table from calculating the true positives divided by sample size. The standard protocol of finding the best one-to-one permutation mapping between learned and ground-truth clusters using linear assignment\cite{kuhn2005hungarian}.
Table 1 are drawn by calculating an evaluation parameter from BSD500. 
\begin{table}
\begin{center}
\begin{tabular}{|l|c|}
\hline
Method & Accuracy \\
\hline\hline
Kmeans\cite{krishna1999genetic} & 0.3639 \\
GS\cite{felzenszwalb2004efficient} & 0.4661 \\
Kanezaki\cite{kanezaki2018unsupervised} & 0.5269\\
Kim\cite{kim2020unsupervised} & 0.5023\\
ours & 0.5384 \\
\hline
 
\end{tabular}
\label{tab111}
\end{center}
\caption{ Comparison pixel accuracy of unsupervised segmentation method in BSD500. }
\end{table}

\section{Conclusions}

\hspace*{1em} In this paper, we proposed a unsupervised image segmentation model based on MMT, and using overlap similarity degree methods for label alignment. The model consists of convolutional filters for feature extraction
and differentiable processes for feature clustering, which enables end-to-end network training.  We have applied this method to unsupervised image segmentation where we achieve comparable performance and drastically improved stable compared to existing methods. 

\hspace*{1em}In future work, we hope to explore the use of other loss functions for
unsupervised image segmentation tasks, such as semantic segmentation loss. We also plan to investigate the use of self-attention mechanism to unsupervised image segmentation.

\bibliographystyle{unsrt}  

{\small

\bibliography{references.bib}

\begin{thebibliography}{10}

\bibitem{pascal-voc-2012}
M.~Everingham, L.~Van~Gool, C.~K.~I. Williams, J.~Winn, and A.~Zisserman.
\newblock The {PASCAL} {V}isual {O}bject {C}lasses {C}hallenge 2012 {(VOC2012)}
  {R}esults.
\newblock
  http://www.pascal-network.org/challenges/VOC/voc2012/workshop/index.html.

\bibitem{martin2001database}
David Martin, Charless Fowlkes, Doron Tal, and Jitendra Malik.
\newblock A database of human segmented natural images and its application to
  evaluating segmentation algorithms and measuring ecological statistics.
\newblock In {\em Proceedings Eighth IEEE International Conference on Computer
  Vision. ICCV 2001}, volume~2, pages 416--423. IEEE, 2001.

\bibitem{krizhevsky2012imagenet}
Alex Krizhevsky, Ilya Sutskever, and Geoffrey~E Hinton.
\newblock Imagenet classification with deep convolutional neural networks.
\newblock In {\em Advances in neural information processing systems}, pages
  1097--1105, 2012.

\bibitem{simonyan2014very}
Karen Simonyan and Andrew Zisserman.
\newblock Very deep convolutional networks for large-scale image recognition.
\newblock {\em arXiv preprint arXiv:1409.1556}, 2014.

\bibitem{long2015fully}
Jonathan Long, Evan Shelhamer, and Trevor Darrell.
\newblock Fully convolutional networks for semantic segmentation.
\newblock In {\em Proceedings of the IEEE conference on computer vision and
  pattern recognition}, pages 3431--3440, 2015.

\bibitem{zheng2015conditional}
Shuai Zheng, Sadeep Jayasumana, Bernardino Romera-Paredes, Vibhav Vineet,
  Zhizhong Su, Dalong Du, Chang Huang, and Philip~HS Torr.
\newblock Conditional random fields as recurrent neural networks.
\newblock In {\em Proceedings of the IEEE international conference on computer
  vision}, pages 1529--1537, 2015.

\bibitem{badrinarayanan2017segnet}
Vijay Badrinarayanan, Alex Kendall, and Roberto Cipolla.
\newblock Segnet: A deep convolutional encoder-decoder architecture for image
  segmentation.
\newblock {\em IEEE transactions on pattern analysis and machine intelligence},
  39(12):2481--2495, 2017.

\bibitem{krishna1999genetic}
K~Krishna and M~Narasimha Murty.
\newblock Genetic k-means algorithm.
\newblock {\em IEEE Transactions on Systems, Man, and Cybernetics, Part B
  (Cybernetics)}, 29(3):433--439, 1999.

\bibitem{felzenszwalb2004efficient}
Pedro~F Felzenszwalb and Daniel~P Huttenlocher.
\newblock Efficient graph-based image segmentation.
\newblock {\em International journal of computer vision}, 59(2):167--181, 2004.

\bibitem{song2020unsupervised}
Liangchen Song, Cheng Wang, Lefei Zhang, Bo~Du, Qian Zhang, Chang Huang, and
  Xinggang Wang.
\newblock Unsupervised domain adaptive re-identification: Theory and practice.
\newblock {\em Pattern Recognition}, 102:107173, 2020.

\bibitem{zhang2019self}
Xinyu Zhang, Jiewei Cao, Chunhua Shen, and Mingyu You.
\newblock Self-training with progressive augmentation for unsupervised
  cross-domain person re-identification.
\newblock In {\em Proceedings of the IEEE International Conference on Computer
  Vision}, pages 8222--8231, 2019.

\bibitem{kanezaki2018unsupervised}
Asako Kanezaki.
\newblock Unsupervised image segmentation by backpropagation.
\newblock In {\em 2018 IEEE international conference on acoustics, speech and
  signal processing (ICASSP)}, pages 1543--1547. IEEE, 2018.

\bibitem{kim2020unsupervised}
Wonjik Kim, Asako Kanezaki, and Masayuki Tanaka.
\newblock Unsupervised learning of image segmentation based on differentiable
  feature clustering.
\newblock {\em IEEE Transactions on Image Processing}, 29:8055--8068, 2020.

\bibitem{ge2020mutual}
Yixiao Ge, Dapeng Chen, and Hongsheng Li.
\newblock Mutual mean-teaching: Pseudo label refinery for unsupervised domain
  adaptation on person re-identification.
\newblock {\em arXiv preprint arXiv:2001.01526}, 2020.

\bibitem{cimpoi2015deep}
Mircea Cimpoi, Subhransu Maji, and Andrea Vedaldi.
\newblock Deep filter banks for texture recognition and segmentation.
\newblock In {\em Proceedings of the IEEE conference on computer vision and
  pattern recognition}, pages 3828--3836, 2015.

\bibitem{He_2019_ICCV}
Junjun He, Zhongying Deng, and Yu~Qiao.
\newblock Dynamic multi-scale filters for semantic segmentation.
\newblock In {\em Proceedings of the IEEE/CVF International Conference on
  Computer Vision (ICCV)}, October 2019.

\bibitem{oquab2014learning}
Maxime Oquab, Leon Bottou, Ivan Laptev, and Josef Sivic.
\newblock Learning and transferring mid-level image representations using
  convolutional neural networks.
\newblock In {\em Proceedings of the IEEE conference on computer vision and
  pattern recognition}, pages 1717--1724, 2014.

\bibitem{cao2018partial}
Zhangjie Cao, Mingsheng Long, Jianmin Wang, and Michael~I Jordan.
\newblock Partial transfer learning with selective adversarial networks.
\newblock In {\em Proceedings of the IEEE Conference on Computer Vision and
  Pattern Recognition}, pages 2724--2732, 2018.

\bibitem{laine2016temporal}
Samuli Laine and Timo Aila.
\newblock Temporal ensembling for semi-supervised learning.
\newblock {\em arXiv preprint arXiv:1610.02242}, 2016.

\bibitem{zhang2018deep}
Ying Zhang, Tao Xiang, Timothy~M Hospedales, and Huchuan Lu.
\newblock Deep mutual learning.
\newblock In {\em Proceedings of the IEEE Conference on Computer Vision and
  Pattern Recognition}, pages 4320--4328, 2018.

\bibitem{fan2012cluster}
Haixiong Fan, Fuxian Liu, and Lu~Xia.
\newblock Cluster label aligning algorithm based on programming model.
\newblock In {\em 2012 24th Chinese Control and Decision Conference (CCDC)},
  pages 1768--1772. IEEE, 2012.

\bibitem{zhou2006clusterer}
Zhi-Hua Zhou and Wei Tang.
\newblock Clusterer ensemble.
\newblock {\em Knowledge-Based Systems}, 19(1):77--83, 2006.

\bibitem{shibata2017misalignment}
Takashi Shibata, Masayuki Tanaka, and Masatoshi Okutomi.
\newblock Misalignment-robust joint filter for cross-modal image pairs.
\newblock In {\em Proceedings of the IEEE International Conference on Computer
  Vision}, pages 3295--3304, 2017.

\bibitem{ji2019invariant}
Xu~Ji, Jo{\~a}o~F Henriques, and Andrea Vedaldi.
\newblock Invariant information clustering for unsupervised image
  classification and segmentation.
\newblock In {\em Proceedings of the IEEE International Conference on Computer
  Vision}, pages 9865--9874, 2019.

\end{thebibliography}
}





\end{document}